\documentclass[a4paper,12pt]{article}
\usepackage[a4paper,margin=1in]{geometry}
\usepackage{graphicx} % Required for inserting images
\usepackage{amsmath}
\usepackage{hyperref}
\usepackage{multicol}
\usepackage{setspace}
\usepackage{animate}

\title{
\textbf{A Generative Approach to High Fidelity 3D Reconstruction from Text Data}
\singlespacing
}

\author{
    \begin{tabular}[t]{c@{\hspace{2cm}} c}
        \textsuperscript{1st} Venkat Kumar R & \textsuperscript{2nd} Deepak Saravanan \\
        {\small \textit{AI \& ML, M.Tech.}} & {\small \textit{Senior 3D Artist}} \\
        {\small \textit{BITS Pilani WILP}} & {\small\textit{India}} \\
           {\small\textit{India}} &  {\small\textit{deepakraj3095cse@gmail.com}} \\
        {\small \textit{venkatkumarr.vk99@gmail.com}}
    \end{tabular}
}

\date{}

\begin{document}

\maketitle
\begin{abstract}
\normalsize
The convergence of generative artificial intelligence and advanced computer vision technologies introduces a groundbreaking approach to transforming textual descriptions into three-dimensional representations. This research proposes a fully automated pipeline that seamlessly integrates text-to-image generation, various image processing techniques, and deep learning methods for reflection removal and 3D reconstruction. By leveraging state-of-the-art generative models like Stable Diffusion, the methodology translates natural language inputs into detailed 3D models through a multi-stage workflow.
\singlespacing
The reconstruction process begins with the generation of high-quality images from textual prompts, followed by enhancement by a reinforcement learning agent and reflection removal using the Stable Delight model. Advanced image upscaling and background removal techniques are then applied to further enhance visual fidelity. These refined two-dimensional representations are subsequently transformed into volumetric 3D models using sophisticated machine learning algorithms, capturing intricate spatial relationships and geometric characteristics. This process achieves a highly structured and detailed output, ensuring that the final 3D models reflect both semantic accuracy and geometric precision.
\singlespacing
This approach addresses key challenges in generative reconstruction, such as maintaining semantic coherence, managing geometric complexity, and preserving detailed visual information. Comprehensive experimental evaluations will assess reconstruction quality, semantic accuracy, and geometric fidelity across diverse domains and varying levels of complexity. By demonstrating the potential of AI-driven 3D reconstruction techniques, this research offers significant implications for fields such as augmented reality (AR), virtual reality (VR), and digital content creation.
\singlespacing
\textbf{Keywords:}
Text-to-3D Reconstruction, Generative AI, Computer Vision, 
Image Generation, Neural Networks, Semantic 3D Modeling
\end{abstract}

\section{Introduction}
\normalsize
The rapid advancement of generative artificial intelligence (AI) and computer vision has opened new possibilities for creating realistic and semantically accurate 3D content from textual descriptions. Traditional 3D content creation often requires specialized skills, significant time investment, and extensive manual intervention, which poses barriers for non-expert users. Generative AI-powered techniques promise to overcome these challenges by automating the 3D modeling process, making it more accessible and efficient.

\singlespacing

This research focuses on developing a fully automated text-to-3D reconstruction pipeline that combines state-of-the-art generative models, image processing techniques, and deep learning algorithms. By transforming natural language prompts into high-quality 3D models, this approach aims to ensure semantic consistency and geometric precision while minimizing the time and expertise required for 3D content development. The proposed methodology enhances visual fidelity through advanced techniques like reinforcement learning-based image enhancement, reflection removal using the Stable Delight model, and sophisticated upscaling and background removal.
\singlespacing
The resulting 3D models find applications across various domains, including gaming, manufacturing, augmented reality (AR), virtual reality (VR), and digital content creation. By democratizing the 3D content generation process, this research seeks to empower a broader audience, enabling creative innovation without the traditional barriers of skill and resource intensity.

\begin{figure}[h]
    \centering
    \includegraphics[width=0.45\textwidth]{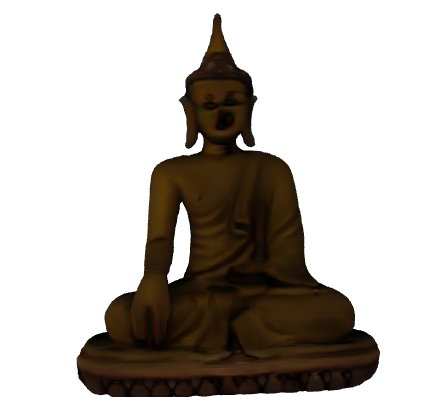}
    \includegraphics[width=0.45\textwidth]{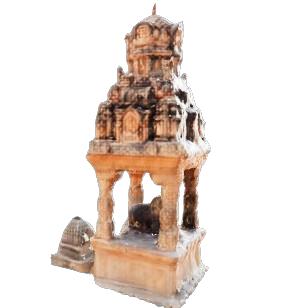}
    \caption{Text-to-3D reconstruction}
    \label{fig:example_images}
\end{figure}

\section{Literature References}

Ian Goodfellow et al. (2014) The seminal GAN paper introduces a framework where a generator and discriminator compete to generate realistic synthetic data, revolutionizing generative modeling across fields. Lvmin Zhang et al. (2023) This paper enhances text-to-image diffusion models by incorporating conditional control, enabling fine-grained visual generation. It improves outputs by introducing additional conditioning methods, such as control over pose, colors, and styles. Christian Ledig et al. ( 2017) SRGAN introduces a GAN-based approach to upscale low-resolution images into high-resolution ones, producing photorealistic details. It uses perceptual loss to capture finer textures that conventional methods fail to achieve. Xuebin Qin et al.(2020) The proposed method of U2-Net introduces a nested U-shaped network architecture designed for efficient and lightweight salient object detection. The model achieves state-of-the-art performance with fewer computational resources. 
\singlespacing
Ming Li et al. (2023) In this ControlNet++ improves conditional diffusion models by incorporating consistency feedback mechanisms, ensuring that the generated output aligns with control inputs like depth maps, edges, and segmentation masks. Alec Radford et al. (2021) CLIP trains a multimodal model on text and image data to learn transferable visual representations. It supports zero-shot image classification and cross-modal retrieval tasks. jiayun Wan et al. (2022) This work proposes a deep learning framework to reconstruct detailed 3D shapes from free-hand 2D sketches. The method bridges artistic input and 3D modeling enabling interactive content creation. Ben Poole et al. (2022) DreamFusion integrates 2D diffusion models with 3D optimization to create realistic 3D models from text prompts. It eliminates the need for 3D-specific training data by leveraging pretrained 2D generative models. H. Lee et al. (2024) This paper uses Neural Radiance Fields (NeRF) to synthesize 3D shapes from text inputs, offering high-quality shape generation. It captures spatial details and scene realism by extending NeRF's capabilities. BERNHARD KERBL et al. (2023) This paper proposes Gaussian splatting to accelerate radiance field rendering, enabling real-time 3D rendering. The method balances computational efficiency with visual quality. Dawid Malarz et al. (2024) Builds on Neural Radiance Fields (NeRF) by using Gaussian splats to represent 3D scenes. It refines color and opacity calculations for improved rendering precision and computational efficiency . Minghua Liu et al. ( 2023) Proposed a framework to generate 3D meshes from a single image in 45 seconds without per-shape optimization. It leverages pre-trained neural networks and templates to ensure high efficiency and quality. 
\singlespacing
I proposed a solution combining state-of-the-art techniques to generate high-quality 3D meshes with detailed texture outputs. This involves text-to-image generation using diffusion models (e.g., CLIP-based or DreamFusion methods) followed by 2D-to-3D reconstruction with fast and geometry-preserving algorithms like One-2-3-45. Advanced GANs and photorealistic super-resolution techniques are integrated for detailed texture mapping. Real-time rendering with Gaussian Splatting ensures efficiency, while semantic controls (e.g., ControlNet++) maintain alignment with the input descriptions. This pipeline generates accurate, textured meshes suitable for gaming, AR/VR, and design applications, making 3D creation more accessible and efficient.

\section{Proposed Method}

The below concept architecture illustrates a generative approach to 3D reconstruction from text and image data, with an emphasis on automating image processing and enhancement using reinforcement-based techniques. It starts with user text input, which is processed through text-to-image generation to create visual representations. These images are then refined using preprocessing techniques like GAN-based upscaling and background removal with U2Net, enhancing quality and isolating subjects. Simultaneously, user-provided images are improved through prompt-based enhancement, incorporating reinforcement learning to adaptively optimize low-light conditions and refine reflections. The preprocessed images are fed into the single-image 3D reconstruction phase, where neural networks perform spatial inference. Finally, the resulting 3D models undergo rigorous evaluation to ensure high-quality outputs suited for real-world applications.

\begin{figure}[h]
    \centering
    \includegraphics[width=0.95\textwidth]{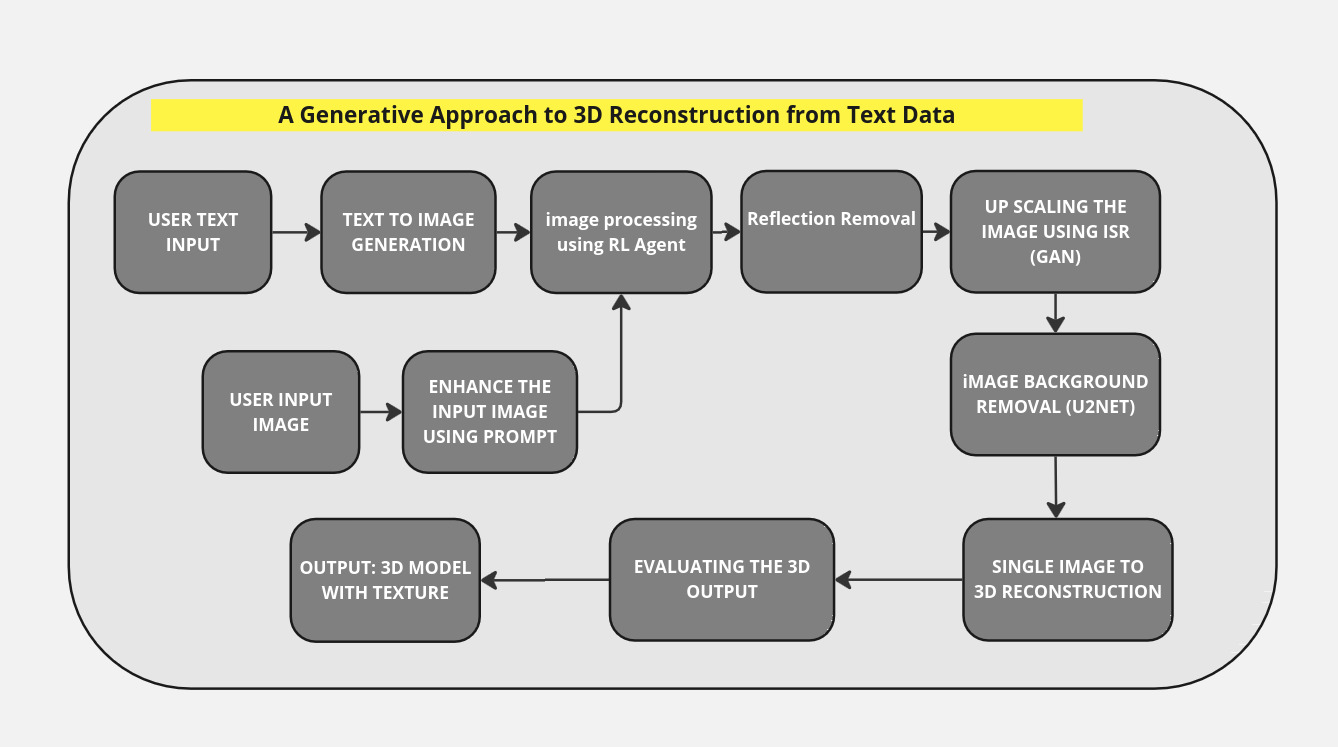}
    \caption{Concept Architecture}
    \label{fig:example_images}
\end{figure}

\subsection{Related work of proposed method}

Podell et al. (2024) introduced SDXL, a state-of-the-art latent diffusion model for high-resolution text-to-image synthesis. By leveraging a threefold larger UNet backbone, additional attention blocks, and a second text encoder, SDXL captures nuanced and contextually rich visual details, setting a new benchmark in generative modeling. Its innovative conditioning schemes and image-to-image refinement techniques ensure exceptional visual fidelity, making SDXL a crucial component for high-quality 3D reconstruction pipelines.

\singlespacing

Tochilkin et al. (2024) presented TripoSR, a fast and efficient 3D object reconstruction model built on transformer architecture. TripoSR generates detailed 3D meshes from a single image in under 0.5 seconds, showcasing remarkable speed and performance. Based on the LRM network architecture, it introduces significant advancements in data processing, model design, and training methodologies. Comprehensive evaluations on public datasets demonstrate TripoSR’s superior quantitative and qualitative performance compared to other open-source solutions, making it a valuable tool for 3D generative AI applications.

\singlespacing

In my pipeline, I incorporate the concepts from both the SDXL text-to-image generation and TripoSR image-to-3D reconstruction models. My approach focuses on enhancing the generated images before the 3D reconstruction phase, refining the visual quality and detail of the images to improve the accuracy and structure of the resulting 3D models. This improvisation ensures that the 3D objects produced exhibit superior structural integrity and intricate detailing, all while maintaining a single-input workflow.

\subsection{Focused area of my approach}

\begin{figure}[h]
    \centering
    \includegraphics[width=0.95\textwidth]{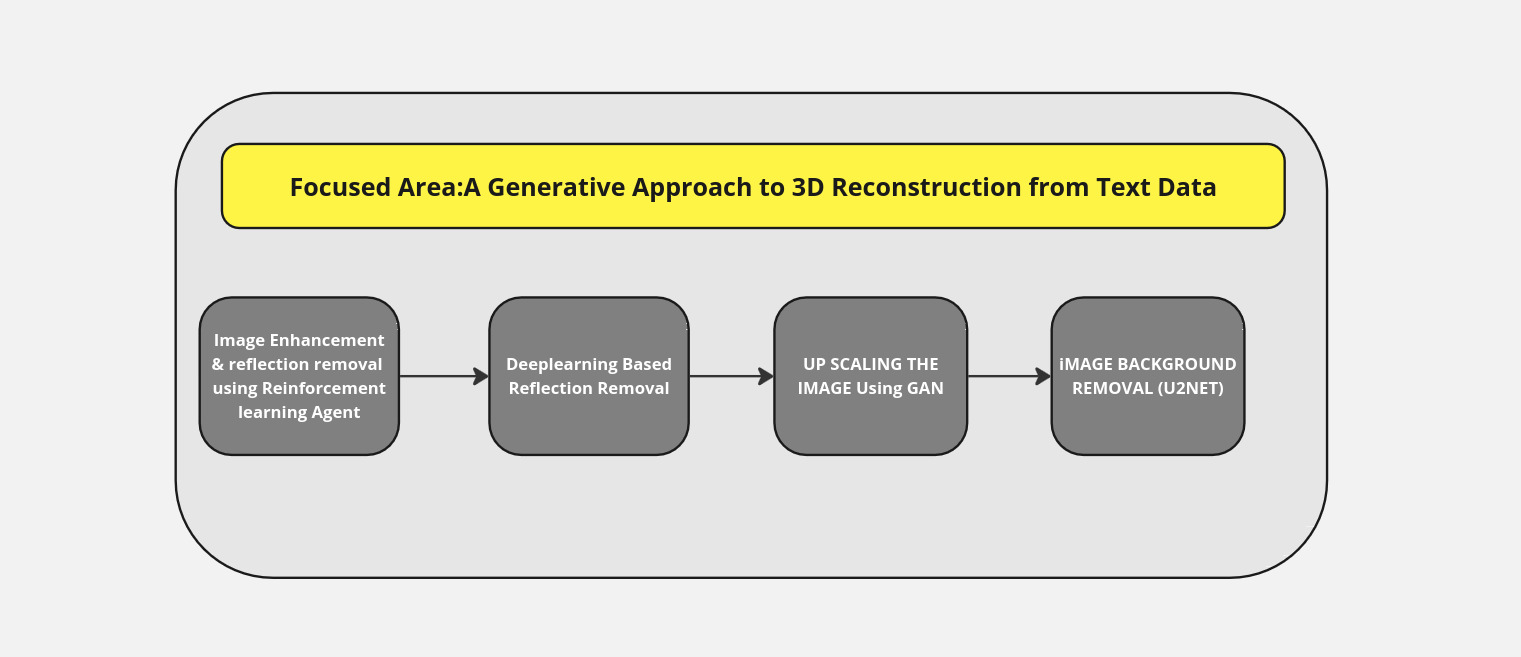}
    \caption{Concept Architecture}
    \label{fig:example_images}
\end{figure}

This paper presents a generative approach to 3D reconstruction from text data, utilizing a sophisticated pipeline that enhances image quality and refines structural detail. The proposed methodology consists of four key stages:

\singlespacing

\textbf{Image Enhancement and Reflection Removal Using a Reinforcement Learning Agent:} The image enhancement process employs a reinforcement learning-based agent, leveraging the Monte Carlo method with deep curve principle-based prediction. This approach optimizes various image attributes, including brightness, smoothness, saturation, contrast, gamma correction, and strength curves. By systematically adjusting these parameters, the model ensures high-quality input data for subsequent stages.

\singlespacing

\textbf{Reflection Removal Using Deep Learning:} For effective reflection removal, we adopt the StableDelight model, a state-of-the-art technique for eliminating specular reflections from textured surfaces. Building upon the stability enhancements introduced in the StableNormal framework, StableDelight applies these advancements to the challenging task of reflection removal. The training dataset comprises Hypersim, Lumos, and various Specular Highlight Removal datasets from TSHRNet. Furthermore, the model incorporates a multi-scale Structural Similarity Index Measure (SSIM) loss and random conditional scales to enhance sharpness and improve one-step diffusion prediction.

\singlespacing

\textbf{Image Upscaling Using Generative Adversarial Networks (GAN):} To achieve high-resolution image outputs, the Super-Resolution GAN (SRGAN) model is employed. This technique generates visually appealing and structurally accurate high-resolution images, which are crucial for the fidelity of the 3D reconstruction process.

\singlespacing

\textbf{Background Removal Using U2Net:} The final stage involves background removal using the U2Net model, an advanced deep learning architecture known for its superior performance in salient object detection. We integrate the REMBG tool to facilitate efficient and precise background segmentation, ensuring that only the relevant foreground information is preserved for 3D reconstruction.

\singlespacing

This comprehensive pipeline significantly enhances the quality and detail of the input data, ultimately leading to more accurate and visually coherent 3D reconstructions. The combination of reinforcement learning, deep learning, and generative techniques underscores the robustness and innovation of our approach.

\section{Experimental Results}

I successfully designed the architecture for the proposed problem, creating a modular pipeline for text-to-image generation, image pre-processing, and image-to-3D reconstruction. The key modules include Stable Diffusion for text-to-image generation, Monte Carlo with a deep curve principle-based prediction (optimizing brightness, smoothness, saturation, contrast, gamma, and strength curve) to determine and apply the best values, Stable Delight for reflection removal, SRGAN for 4x image upscaling, and neural networks for 3D reconstruction. This approach ensured smooth data flow, clear dependencies, and effective integration of models and frameworks.

\begin{figure}[h]
    \centering
    \includegraphics[width=0.45\textwidth]{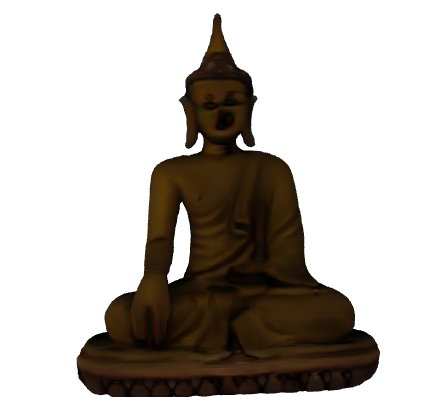}
    \includegraphics[width=0.45\textwidth]{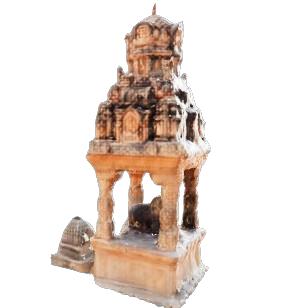}
    \caption{Text-to-3D reconstruction}
    \label{fig:example_images}
\end{figure}

\doublespacing

\textbf{The images below represent text-to-3D reconstruction and single-image-to-3D reconstruction, showcasing all views of the 3D models.}

\doublespacing

\begin{figure}[h]
    \centering
    \includegraphics[width=0.45\textwidth]{Images/1.png}
    \includegraphics[width=0.45\textwidth]{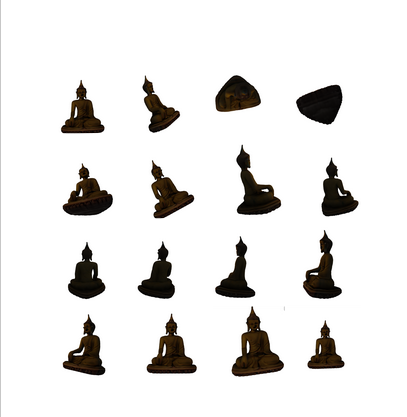}
    \caption{Text to 3D: 3D Reconstruction and All View Output}
    \includegraphics[width=0.45\textwidth]{Images/1_1.png}
    \includegraphics[width=0.45\textwidth]{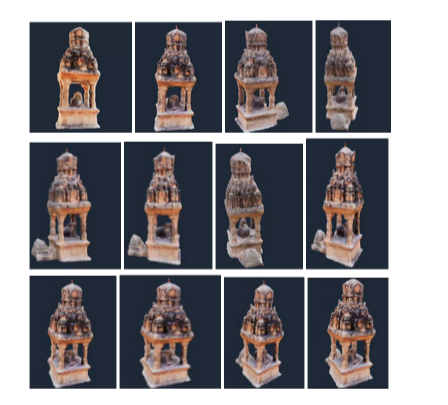}
    \caption{Single Image to 3D: 3D Reconstruction and All View Output}
    \label{fig:example_images}
\end{figure}

\section{Future Developement}
Our research trajectory encompasses several promising avenues for advancing our current approach. Primarily, we intend to develop a sophisticated reinforcement learning agent specifically designed for texture and color optimization. This agent will enable more nuanced and aesthetically cohesive results by dynamically learning and adapting color and texture parameters.
\singlespacing
A key focus will be integrating material detection with base color application techniques. This integration will facilitate the generation of more photorealistic textures and support more granular surface refinements. By establishing a more sophisticated relationship between material characteristics and color mapping, we anticipate significant improvements in surface rendering quality.
\singlespacing
Furthermore, we are committed to developing an advanced UV mapping model that will enhance texture alignment and projection onto three-dimensional surfaces. This model will address current limitations in surface mapping precision, ensuring more accurate and visually seamless texture representation.
\singlespacing
The overarching objective of these strategic developments is to achieve high-quality text-to-3D reconstruction while simultaneously reducing computational overhead. By focusing on efficiency and intelligent learning mechanisms, we aim to establish a more streamlined and resource-effective approach compared to existing methodologies.

\section{Performance and Evaluation}
Currently, formal evaluation metrics such as F-score and Chamfer distance have not been implemented. Instead, the quality of the 3D reconstruction results has been assessed through visual inspection and standard analysis conducted by a professional 3D artist. The feedback received indicated that the approach produces promising results, with suggestions to further improve the tripod structure. Future work will focus on integrating quantitative evaluation metrics to provide a more comprehensive assessment of model performance.

\section{Conclusion}
This generative approach to 3D reconstruction from text and image data streamlines the automation of image processing and enhancement. By integrating text-to-image generation,GAN-based upscaling, background removal, and reinforcement learning-based optimizations, the system ensures high-quality visual representations. The single-image 3D reconstruction phase effectively infers spatial structures, producing detailed 3D models that align with real-world applications. The methodology demonstrates a robust pipeline for generating accurate and high-fidelity reconstructions.

\newpage
\section{Acknowledgment}
I extend my gratitude to all the contributors and researchers who have made the Diffusion model, Stable Diffusion, and TripoSR models accessible, enabling groundbreaking advancements in my approaches. I believe my framework provides more accurate structural details and enhances the quality of the 3D output. I’m happy to share these insights with everyone.

\newpage
\section{Reference}
Ian Goodfellow et al. (2014) The seminal GAN paper introduces a framework where a generator and discriminator compete to generate realistic synthetic data, revolutionizing generative modeling across fields.
\singlespacing
Christian Ledig et al. ( 2017) SRGAN introduces a GAN-based approach to upscale low-resolution images into high-resolution ones, producing photorealistic details. It uses perceptual loss to capture finer textures that conventional methods fail to achieve. 
\singlespacing
Xuebin Qin et al. (2020)The proposed method of U2-Net introduces a nested U-shaped network architecture designed for efficient and lightweight salient object detection. The model achieves
state-of-the-art performance with fewer computational resources.
\singlespacing
Alec Radford et al. (2021) CLIP trains a multimodal model on text and image data to learn transferable visual representations. It supports zero-shot image classification and cross-modal retrieval tasks.
\singlespacing
jiayun Wan et al. (2022) This work proposes a deep learning framework to reconstruct detailed 3D shapes from free-hand 2D sketches. The method bridges artistic input and 3D modeling enabling interactive content creation. 
\singlespacing
Ben Poole et al. (2022) DreamFusion integrates 2D diffusion models with 3D optimization to create realistic 3D models from text prompts. It eliminates the need for 3D-specific training data by leveraging pretrained 2D generative models. 
\singlespacing
Lvmin Zhang et al. (2023) This paper enhances text-to-image diffusion models by incorporating conditional control, enabling fine-grained visual generation. It improves outputs by introducing additional conditioning methods, such as control over pose, colors, and styles. 
\singlespacing
BERNHARD KERBL et al. (2023) This paper proposes Gaussian splatting to accelerate radiance field rendering, enabling real-time 3D rendering. The method balances computational efficiency with visual quality. 
\singlespacing
Dawid Malarz et al. (2024) Builds on Neural Radiance Fields (NeRF) by using Gaussian splats to represent 3D scenes. It refines color and opacity calculations for improved rendering precision and computational efficiency . 
\singlespacing
Minghua Liu et al. ( 2024) Proposed a framework to generate 3D meshes from a single image in 45 seconds without per-shape optimization. It leverages pre-trained neural networks and templates to ensure high efficiency and quality. 
\singlespacing
StableDelight: Revealing Hidden Textures by Removing Specular Reflections
\singlespacing
Reinforcement learning Book by Sutton Barto

\end{document}